\RequirePackage{silence}
\documentclass[journal,twoside,web]{ieeecolor}
\usepackage{generic}
\usepackage{cite}
\usepackage{amsmath,amssymb,amsfonts}
\usepackage{graphicx}
\usepackage[hidelinks]{hyperref}
\usepackage{textcomp}

\usepackage{booktabs}
\usepackage{multirow}
\usepackage[acronym]{glossaries}
\glsdisablehyper

\newacronym{ehr}{EHR}{electronic health record}
\newacronym{icu}{ICU}{intensive care unit}
\newacronym{vae}{VAE}{variational auto-encoder}
\newacronym{gan}{GAN}{generative adversarial network}
\newacronym{ddpm}{DDPM}{denoising diffusion probabilistic model}
\newacronym{sde}{SDE}{stochastic differential equation}
\newacronym{bigru}{BiGRU}{bidirectional gated recurrent unit}
\newacronym{cdtd}{CDTD}{continuous diffusion model for mixed-type tabular data}
\newacronym{cfg}{CFG}{classifier-free guidance}
\newacronym{ode}{ODE}{ordinary differential equation}
\newacronym{tstr}{TSTR}{train on synthetic, test on real}
\newacronym{trtr}{TRTR}{train on real, test on real}
\newacronym{mmd}{MMD}{maximum mean discrepancy}
\newacronym{corr-mae}{Corr MAE}{correlation matrix mean absolute error}
\newacronym{acf-mse}{ACF MSE}{autocorrelation function mean squared error}
\newacronym{dtw}{DTW}{dynamic time warping}
\newacronym{tvd}{TVD}{total variation distance}
\newacronym{trans-dist}{Trans Dist}{transition matrix distance}
\newacronym{c2st}{C2ST}{classifier two-sample test}
\newacronym{lstm}{LSTM}{long short-term memory}
\newacronym{auc}{AUC}{area under the receiver operating characteristic curve}
\newacronym{mlp}{MLP}{multi-layer perceptron}

\def\BibTeX{{\rm B\kern-.05em{\sc i\kern-.025em b}\kern-.08em
    T\kern-.1667em\lower.7ex\hbox{E}\kern-.125emX}}
\begin{document}
\title{CDMT-EHR: A Continuous-Time Diffusion Framework for Generating Mixed-Type Time-Series Electronic Health Records}
\author{Shaonan Liu, Yuichiro Iwashita, Soichiro Nakako, Masakazu Iwamura, \IEEEmembership{Senior Member, IEEE}, and Koichi Kise, \IEEEmembership{Member, IEEE}
\thanks{This research was supported in part by JST ASPIRE (\mbox{JPMJAP2403}), and JSPS Grant-in-Aid for Challenging Research (Exploratory) (24K22339). Code will be made available upon acceptance. (Corresponding author: Yuichiro Iwashita.)}
\thanks{Shaonan Liu, Masakazu Iwamura, and Koichi Kise are with the Department of Core Informatics, Graduate School of Informatics, Osaka Metropolitan University, Sakai, Osaka 599-8531, Japan (e-mail: sc24697f@st.omu.ac.jp; masa.i@omu.ac.jp; kise@omu.ac.jp).}
\thanks{Yuichiro Iwashita is with the Smart Data \& Knowledge Services Department, German Research Center for Artificial Intelligence (DFKI GmbH), 67663 Kaiserslautern, Germany, and the Department of Computer Science, RPTU University Kaiserslautern-Landau, 67663 Kaiserslautern, Germany (e-mail: yuichiro.iwashita@dfki.de).}
\thanks{Soichiro Nakako is with the Department of Hematology and Department of Laboratory Medicine and Medical Informatics, Graduate School of Medicine, Osaka Metropolitan University, Osaka 545-8585, Japan (e-mail: s-nakako@omu.ac.jp).}}

\maketitle

\begin{abstract}
Electronic health records (EHRs) are invaluable for clinical research, yet privacy concerns severely restrict data sharing. Synthetic data generation offers a promising solution, but EHRs present unique challenges: they contain both numerical and categorical features that evolve over time. While diffusion models have demonstrated strong performance in EHR synthesis, existing approaches predominantly rely on discrete-time formulations, which suffer from finite-step approximation errors and coupled training-sampling step counts. We propose a continuous-time diffusion framework for generating mixed-type time-series EHRs with three contributions: (1) continuous-time diffusion with a bidirectional gated recurrent unit backbone for capturing temporal dependencies, (2) unified Gaussian diffusion via learnable continuous embeddings for categorical variables, enabling joint cross-feature modeling, and (3) a factorized learnable noise schedule that adapts to per-feature-per-timestep learning difficulties. Experiments on two large-scale intensive care unit datasets demonstrate that our method outperforms existing approaches in downstream task performance, distribution fidelity, and discriminability, while requiring only 50 sampling steps compared to 1,000 for baseline methods. Classifier-free guidance further enables effective conditional generation for class-imbalanced clinical scenarios.
\end{abstract}

\begin{IEEEkeywords}
Continuous-time diffusion, electronic health records, synthetic data.
\end{IEEEkeywords}

\glsresetall

\section{Introduction}
\label{sec:introduction}

\IEEEPARstart{E}{lectronic} health records (EHRs\glsunset{ehr}) contain a wealth of clinical information, including vital signs and laboratory measurements, and play a critical role in clinical research. With advances in machine learning, \glspl{ehr} have been increasingly applied to downstream tasks such as disease diagnosis and in-hospital mortality prediction, underscoring their growing importance~\cite{shickel2017deep}. However, the sensitive personal information embedded in \glspl{ehr} significantly restricts data access and sharing, which hinders the development of machine learning models for healthcare applications.

Several approaches have been proposed to address this challenge, but each has limitations. Anonymization techniques such as data masking have been widely adopted, yet re-identification risks remain high~\cite{benitez2010evaluating}, and such processing can distort the statistical properties of the original data.

Against this background, generating synthetic data that preserves the statistical characteristics of real \glspl{ehr} has emerged as a promising alternative. \Gls{ehr} synthesis must simultaneously address two key challenges. (1) \Glspl{ehr} contain a mixture of numerical variables (e.g., body temperature and blood pressure) and categorical variables (e.g., diagnostic codes and medication flags), which requires a unified framework to model heterogeneous data types. (2) \Glspl{ehr} are time-series data in which patient states evolve over time, which necessitates models that capture temporal dependencies. Various generative models have been proposed, including rule-based~\cite{franklin2014plasmode, walonoski2018synthea}, \gls{vae}-based~\cite{biswal2021eva, lee2024leveraging}, and \gls{gan}-based~\cite{choi2017generating, baowaly2019synthesizing, torfi2020corgan, yoon2023ehr, karami2025timehr} methods. However, these approaches have not adequately addressed both challenges simultaneously.

Recently, diffusion models~\cite{ho2020denoising, song2021scorebased}, which have demonstrated superior performance in image generation, have attracted attention for \gls{ehr} synthesis. Diffusion models generate high-quality data by progressively adding noise in a forward diffusion process and then learning to reverse this process through a denoising model. Several diffusion-based \gls{ehr} generation methods have been proposed~\cite{he2023meddiff, ceritli2023synthesizing, naseer2023scoehr, yuan2023ehrdiff, lee2023codi, han2025guided, deng2025tardiff}. However, most focus on either a single variable type or non-temporal data, leaving the mixed-type time-series challenge largely unaddressed. TimeDiff~\cite{tian2024reliable}, which uses a bidirectional RNN-based discrete-time diffusion model for mixed-type temporal \glspl{ehr}, represents a notable exception. Nevertheless, TimeDiff and most other diffusion-based \gls{ehr} methods rely on \glspl{ddpm}, which have the following inherent limitations~\cite{song2021scorebased}. (1) The forward process is approximated by finite discrete steps, which introduces approximation errors. (2) The sampling step count is coupled with the training step count, which degrades sampling efficiency. (3) A fixed noise schedule shared across all features cannot adapt to the heterogeneous marginal distributions present in \glspl{ehr}.

Continuous-time diffusion models~\cite{song2021scorebased} address the first two limitations by formulating the diffusion process using \glspl{sde}, which decouple training and sampling step counts. For mixed-type static tabular data, continuous-time diffusion frameworks such as TabDiff~\cite{shi2025TabDiff} and the \gls{cdtd}~\cite{mueller2025continuous} have demonstrated state-of-the-art performance. TabDiff addresses the schedule limitation through learnable per-feature noise schedules and employs separate diffusion processes for numerical (Gaussian) and categorical (mask diffusion) features, while \gls{cdtd} maps categorical variables into a continuous embedding space to apply unified Gaussian diffusion. However, neither framework supports time-series data, leaving a gap for temporal \gls{ehr} generation.

In this work, we propose CDMT-EHR, a continuous-time diffusion framework for mixed-type time-series \gls{ehr} generation by extending TabDiff with the following three contributions:
\begin{enumerate}
    \item \textbf{Continuous-time diffusion for temporal \gls{ehr} generation:} We introduce continuous-time diffusion to mixed-type time-series \gls{ehr} generation by incorporating a \gls{bigru}~\cite{cho2014learning} as the denoising backbone. This enables flexible sampling with decoupled training and sampling step counts while capturing temporal dependencies through the hidden states of the \gls{bigru}.

    \item \textbf{Unified Gaussian diffusion via continuous embeddings:} Following \gls{cdtd}~\cite{mueller2025continuous}, we map categorical variables into a learnable continuous embedding space and apply the same Gaussian diffusion process as for numerical variables. This unified continuous-space diffusion enables more effective joint modeling of cross-feature dependencies compared to combining separate Gaussian and mask diffusion processes.

    \item \textbf{Factorized learnable noise schedule:} We propose a factorized learnable noise schedule that decomposes each feature-timestep pair into global, feature-specific, and time-specific components. This extends TabDiff's per-feature schedule to the temporal setting while reducing the number of learnable parameters from $O(F \times L)$ to $O(F + L)$, where $F$ is the number of features and $L$ is the sequence length, adapting to per-feature-per-timestep learning difficulties.
\end{enumerate}

We evaluate our method on two large-scale \gls{icu} datasets, including MIMIC-III~\cite{johnson_pollard_mark_2016_mimic,johnson_et_al_2016_mimic_iii,goldberger_et_al_2000_physionet} and eICU~\cite{pollard2018eicu}, across three perspectives, namely downstream task performance, fidelity, and discriminability. Our method consistently outperforms the existing TimeDiff baseline while requiring only 50 sampling steps compared to 1,000 steps for TimeDiff. Furthermore, we demonstrate that \gls{cfg} enables effective conditional generation for data augmentation in class-imbalanced clinical scenarios.

\section{Related Work}
\label{sec:related_work}

\subsection{\texorpdfstring{Non-Diffusion \gls{ehr} Generation Models}{Non-Diffusion EHR Generation Models}}

Early approaches to synthetic \gls{ehr} generation include rule-based methods~\cite{franklin2014plasmode, walonoski2018synthea}, which generate data according to predefined clinical rules but produce limited diversity. With the development of deep generative models, \gls{vae}-based methods~\cite{biswal2021eva, lee2024leveraging} offer stable training but often yield lower-quality samples, while \gls{gan}-based methods~\cite{choi2017generating, baowaly2019synthesizing, torfi2020corgan, yoon2023ehr, karami2025timehr} can produce high-quality data but suffer from mode collapse and training instability~\cite{yuan2023ehrdiff}. Furthermore, most existing studies target simplified data structures, e.g., only diagnostic codes, that do not fully reflect the complexity of real-world \glspl{ehr}.

\subsection{\texorpdfstring{Diffusion-Based \gls{ehr} Generation Models}{Diffusion-Based EHR Generation Models}}

Diffusion models have been increasingly applied to \gls{ehr} generation~\cite{he2023meddiff, ceritli2023synthesizing, naseer2023scoehr, yuan2023ehrdiff, lee2023codi, han2025guided, zhong2024synthesizing, he2024flexible, deng2025tardiff}, demonstrating improvements in both data quality and privacy over conventional generative models. However, most of these methods are limited to a single variable type or non-temporal data. TimeDiff~\cite{tian2024reliable} addresses this gap by applying a \gls{ddpm}-based diffusion model with a bidirectional RNN backbone to mixed-type temporal \glspl{ehr}. While TimeDiff achieves competitive performance, it inherits the limitations of discrete-time diffusion, including finite-step approximation errors, coupled training-sampling step counts, and a fixed noise schedule that cannot adapt to heterogeneous feature distributions.

\subsection{Continuous-Time Diffusion for Mixed-Type Data}

To overcome the limitations of discrete-time diffusion, TabDiff~\cite{shi2025TabDiff} proposes continuous-time diffusion frameworks for mixed-type static tabular data. TabDiff uses Gaussian diffusion for numerical features and mask diffusion~\cite{sahoo2024simple} for categorical features, with learnable per-feature noise schedules. However, mask diffusion constrains categorical representations to either a fixed category or a mask state, which prevents smooth inter-category transitions and potentially limits generation quality for temporal data where categories may evolve over time.

\Gls{cdtd}~\cite{mueller2025continuous} addresses this by mapping categorical variables into a continuous embedding space and applying unified Gaussian diffusion across all features. This enables joint modeling of cross-feature dependencies in a shared continuous space. However, \gls{cdtd} applies a common learnable noise schedule per variable type rather than per individual feature, which may not optimally adapt to features with diverse marginal distributions within the same type.

Critically, both TabDiff and \gls{cdtd} are designed for static (non-temporal) tabular data. Our work extends TabDiff to temporal sequences while incorporating the continuous embedding approach of \gls{cdtd} and introducing a factorized per-feature-per-timestep noise schedule.

\section{Proposed Method}
\label{sec:method}

\begin{figure*}[tbp]
    \centering
    \includegraphics[width=\textwidth]{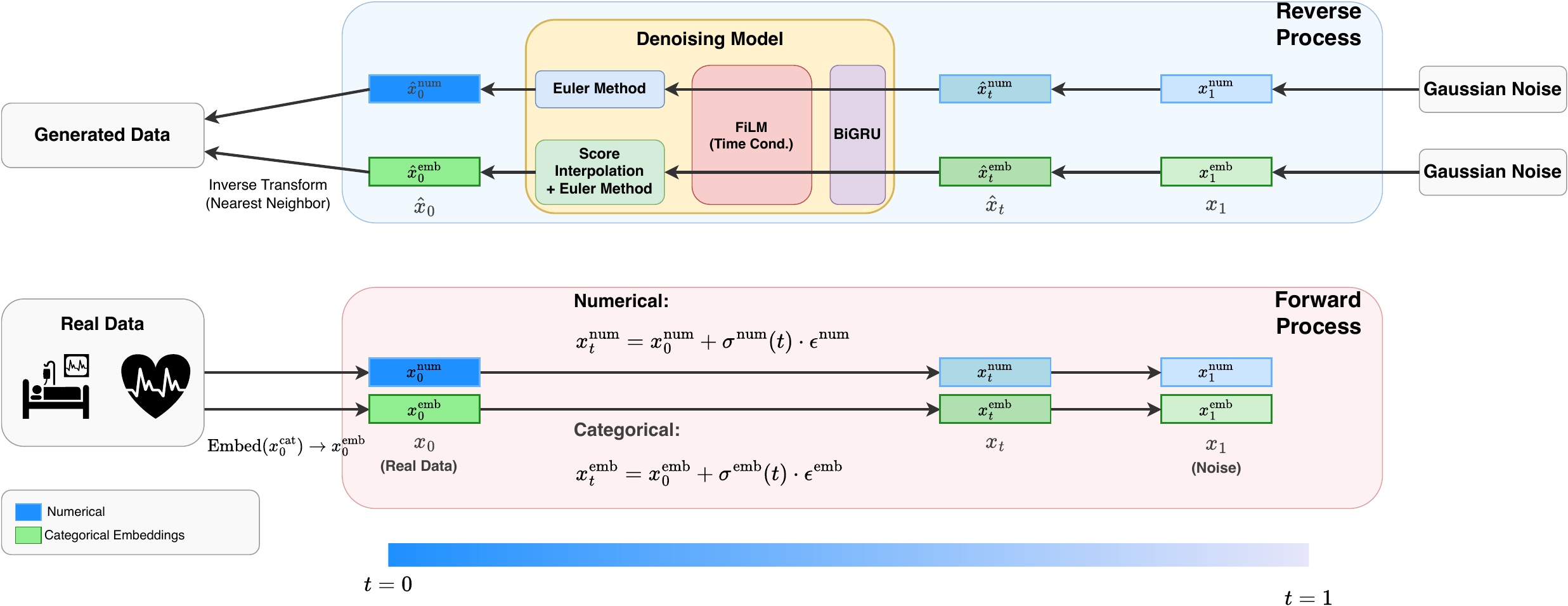}
    \caption{Overview of the proposed method. Categorical variables are mapped into a learnable continuous embedding space, and both numerical and categorical (embedded) features undergo unified Gaussian diffusion. A \gls{bigru} backbone captures temporal dependencies. Factorized learnable noise schedules adapt to per-feature-per-timestep learning difficulties.}
    \label{fig:overview}
\end{figure*}

We propose a continuous-time diffusion framework for mixed-type time-series \gls{ehr} generation that extends TabDiff~\cite{shi2025TabDiff} through three key modifications: (1) a temporal extension using a \gls{bigru} backbone for efficient temporal modeling with label information propagation, (2) unified Gaussian diffusion via continuous embeddings for categorical variables, and (3) factorized learnable noise schedules. Fig.~\ref{fig:overview} presents an overview of the proposed method. In the following subsections, we first introduce the necessary preliminaries from TabDiff, then describe each contribution in detail.

\subsection{Preliminaries: TabDiff}
\label{sec:preliminaries}

TabDiff~\cite{shi2025TabDiff} is a continuous-time diffusion model for mixed-type static tabular data. An individual data sample $x$ is represented as the concatenation of numerical and categorical features: $x = [x^{\mathrm{num}}, x^{\mathrm{cat}}]$, where $x^{\mathrm{num}} \in \mathbb{R}^{M_{\mathrm{num}}}$ denotes $M_{\mathrm{num}}$-dimensional numerical features and $x^{\mathrm{cat}}$ denotes $M_{\mathrm{cat}}$ categorical features, with the $j$-th categorical feature represented as a one-hot vector $(x^{\mathrm{cat}})_j \in \{0,1\}^{C_j+1}$, where $C_j$ is the number of categories and the additional dimension represents a special mask state.

TabDiff formulates the diffusion process in continuous time $t \in [0,1]$, where $t=0$ corresponds to clean data and $t=1$ to pure noise. For numerical features, a variance-exploding \gls{sde}~\cite{song2021scorebased, karras2022elucidating} defines the forward process:
\begin{align}
    \label{eq:gaussian_forward}
    x_t^{\mathrm{num}} = x_0^{\mathrm{num}} + \sigma^{\mathrm{num}}(t)\,\epsilon,
    \quad \epsilon \sim \mathcal{N}(0, I_{M_{\mathrm{num}}})
\end{align}
where $\sigma^{\mathrm{num}}(t)$ is the noise schedule for numerical features.
The corresponding reverse process is given by an \gls{ode}:
\begin{align}
    \label{eq:gaussian_reverse}
    dx^{\mathrm{num}} = -\left[\frac{d}{dt}\sigma^{\mathrm{num}}(t)\right]\sigma^{\mathrm{num}}(t)\,
    \nabla_x \log p_t(x^{\mathrm{num}})\,dt
\end{align}

TabDiff introduces learnable per-feature noise schedules using a power-mean
parameterization. For the $i$-th numerical feature:
\begin{align}
    \label{eq:power_mean}
    \sigma_{\rho_i}^{\mathrm{num}}(t) =
    (\sigma_{\min}^{1/\rho_i} +
    t (\sigma_{\max}^{1/\rho_i} - \sigma_{\min}^{1/\rho_i})
    )^{\rho_i}
\end{align}
where $\rho_i$ is a learnable parameter for the $i$-th feature, and
$\sigma_{\min}$ and $\sigma_{\max}$ are fixed hyperparameters that anchor the
noise range.

For categorical features, TabDiff uses mask diffusion~\cite{sahoo2024simple}, which progressively transitions categories to a special mask state. However, as discussed in Section~\ref{sec:related_work}, this approach has limitations for temporal data.

A denoising model $f_\theta$, which approximates the score function $\nabla_x \log p_t(x)$ in \eqref{eq:gaussian_reverse}, is jointly trained on both feature types, and TabDiff's total loss combines a noise prediction loss for numerical features and a masked cross-entropy loss for categorical features.

\subsection{\texorpdfstring{Temporal Extension with \gls{bigru}}{Temporal Extension with BiGRU}}
\label{sec:temporal_extension}

TabDiff processes each data sample as a single row of a table with shape $[N, F]$, where $N$ is the number of samples and $F$ is the number of features. Real-world \glspl{ehr}, however, are time-series with shape $[N, F, L]$, where $L$ is the number of time steps. Directly extending TabDiff's Transformer backbone to the temporal dimension would incur prohibitive attention complexity, and Transformer-based alternatives such as TimeSformer~\cite{bertasius2021space} limit label information propagation across features due to their separate temporal and spatial attention design.

We therefore adopt a \gls{bigru}~\cite{cho2014learning} as the denoising backbone, following TimeDiff~\cite{tian2024reliable}. At each time step, the \gls{bigru} receives a feature vector concatenating all variables (including labels for conditional generation; see Section~\ref{sec:cfg}), allowing label information to naturally propagate via hidden states with $O(F \cdot L)$ complexity.

Within the \gls{bigru} backbone, we apply layer normalization followed by feature-wise linear modulation for diffusion time conditioning:
\begin{align}
    \label{eq:film}
    h_{\mathrm{out}} = h_{\mathrm{in}} \cdot (1 + \gamma(t)) + \omega(t)
\end{align}
where $h_{\mathrm{in}}$ and $h_{\mathrm{out}}$ are the input and output latent representations within the \gls{bigru}, and $\gamma(t)$ and $\omega(t)$ are scale and shift parameters learned from the embedding of the diffusion time step $t$. This enables the model to adapt its behavior across different noise levels.

\subsection{Unified Gaussian Diffusion via Continuous Embeddings}
\label{sec:continuous_embedding}

As discussed in Sections~\ref{sec:related_work} and~\ref{sec:preliminaries}, TabDiff's mask diffusion for categorical features prevents smooth intermediate representations and hinders unified cross-feature modeling. To address these limitations, we adopt the continuous embedding approach of \gls{cdtd}~\cite{mueller2025continuous}. For the $j$-th categorical feature with $C_j$ categories, each category $k \in \{0, 1, \ldots, C_j - 1\}$ is assigned a learnable $d$-dimensional embedding vector $e_{j,k} \in \mathbb{R}^d$. The embedding is L2-normalized and scaled:
\begin{align}
    \label{eq:embedding}
    x^{\mathrm{emb}} = \frac{e_{j,k}}{\|e_{j,k}\|_2} \cdot \sqrt{d}
\end{align}

The same Gaussian forward process \eqref{eq:gaussian_forward} and reverse \gls{ode} \eqref{eq:gaussian_reverse} are then applied to these continuous embeddings, with a separate noise schedule $\sigma^{\mathrm{emb}}(t)$.

By applying unified Gaussian diffusion to both numerical features and categorical embeddings, all features diffuse in a shared continuous space. This enables the model to jointly learn dependencies between numerical and categorical features more effectively than when separate diffusion processes are used.

\subsection{Factorized Learnable Noise Schedules}
\label{sec:factorized_schedule}

\Gls{ehr} features exhibit heterogeneous marginal distributions; body temperature and blood pressure, for instance, have very different statistical characteristics, and the properties of each feature may also vary across time steps due to missing data patterns and temporal dynamics. This heterogeneity leads to uneven learning difficulty across features and time steps.

A naive approach of assigning an independent $\rho$ parameter to each feature-timestep pair would require $O(F \times L)$ parameters, raising scalability and overfitting concerns. Instead, we propose a factorized parameterization that decomposes $\rho$ into additive components:
\begin{align}
    \label{eq:factorized_rho}
    \rho_{f,l} = \rho_{\mathrm{global}} +
    \rho_{\mathrm{feature}}[f] + \rho_{\mathrm{time}}[l]
\end{align}
where $\rho_{\mathrm{global}}$ is a scalar shared across all features and time steps, $\rho_{\mathrm{feature}}[f]$ is an adjustment for feature $f$, and $\rho_{\mathrm{time}}[l]$ is an adjustment for time step $l$. Both $\rho_{\mathrm{feature}}$ and $\rho_{\mathrm{time}}$ are initialized to zero, so that learning proceeds from the global baseline. This $\rho_{f,l}$ is substituted into \eqref{eq:power_mean} in place of $\rho_i$, yielding per-feature-per-timestep noise schedules with only $O(F + L)$ parameters.

Since categorical features now undergo Gaussian diffusion in the continuous embedding space (Section~\ref{sec:continuous_embedding}), the same power-mean schedule can be applied to both feature types. We apply separate power-mean schedules for numerical and categorical (embedded) features, each with its own set of factorized parameters. This allows different $\sigma_{\max}$ values for each type: we use
$\sigma_{\max}^{\mathrm{num}} = 80$ for numerical features and
$\sigma_{\max}^{\mathrm{emb}} = 100$ for categorical embeddings, following
\gls{cdtd}~\cite{mueller2025continuous}. The larger $\sigma_{\max}$ for embeddings
accounts for the fact that the L2-normalized embeddings have norm $\sqrt{d}$
after \eqref{eq:embedding}, requiring a greater noise scale for full diffusion.

\subsection{Loss Function}
\label{sec:loss}

The change in categorical diffusion from mask-based to Gaussian embedding-based necessitates a corresponding change in the loss function. The categorical loss of TabDiff computes a masked cross-entropy only at positions in the mask state, which is not applicable when mask states do not exist.

\subsubsection{Numerical loss}
For numerical features, we adopt a data-prediction formulation based on
EDM~\cite{karras2022elucidating}, where the denoising model directly predicts
the clean data $\hat{x}_0^{\mathrm{num}}$ rather than the noise $\epsilon$.
Data prediction is mathematically equivalent to noise
prediction~\cite{karras2022elucidating} but is more naturally aligned with
the categorical loss design, where the model predicts category probabilities
from noisy embeddings. The numerical loss applies a noise-level-dependent
weight $\lambda(\sigma)$, where $\sigma = \sigma^{\mathrm{num}}(t)$, for training
stability:
\begin{align}
    \label{eq:num_loss}
    \mathcal{L}_{\text{num}}(\theta, \rho) = \mathbb{E}_{x_0, t, \epsilon}
    \left[\lambda(\sigma) \cdot
    \|\hat{x}_0^{\mathrm{num}} - x_0^{\mathrm{num}}\|_2^2\right]
\end{align}

\subsubsection{Categorical loss}
For categorical features, following \gls{cdtd}~\cite{mueller2025continuous}, we use
a cross-entropy loss between the model's predicted category probabilities and
the ground-truth categories. Unlike the mask diffusion loss, this loss is
computed at \emph{all} positions:
\begin{align}
    \label{eq:cat_loss}
    \mathcal{L}_{\text{emb}}(\theta, \rho) &= \mathbb{E}_{x_0, t, \epsilon} \Biggl[ -\frac{1}{M_{\mathrm{cat}} L} \sum_{j,l} \log p_\theta ( x_0^{\mathrm{cat},(l,j)} \mid x_t^{\mathrm{emb}}, t ) \Biggr]
\end{align}
where $x_0^{\mathrm{cat},(l,j)}$ is the ground-truth category of the $j$-th
categorical feature at time step $l$, and
$p_\theta(\cdot \mid x_t^{\mathrm{emb}}, t)$ is the category probability
computed via softmax from the model's logit output.

\subsubsection{Total loss}
The two losses are combined with weighting coefficients
$\lambda_{\mathrm{num}}$ and $\lambda_{\mathrm{emb}}$:
\begin{align}
    \label{eq:total_loss}
    \mathcal{L}(\theta, \rho) =
    \lambda_{\mathrm{num}} \mathcal{L}_{\text{num}}(\theta, \rho) +
    \lambda_{\mathrm{emb}} \mathcal{L}_{\text{emb}}(\theta, \rho)
\end{align}

\subsection{Training and Sampling}
\label{sec:training_sampling}

The training and sampling procedures are summarized in Algorithms~1 and~2 in the supplementary material.

During training, the denoising model uses the EDM~\cite{karras2022elucidating} preconditioning framework to normalize input and output scales across noise levels.

During sampling, the reverse \gls{ode} \eqref{eq:gaussian_reverse} is solved numerically using the Euler method. For numerical features, the score function is approximated as $\nabla_x \log p_t(x^{\mathrm{num}}) \approx (x_t^{\mathrm{num}} - \hat{x}_0^{\mathrm{num}}) / (\sigma^{\mathrm{num}}(t))^2$ and substituted into \eqref{eq:gaussian_reverse} to derive the Euler update. For categorical features, the denoised embedding estimate $\hat{x}_0^{\mathrm{emb}}$ is computed via score interpolation~\cite{mueller2025continuous}:
\begin{align}
    \label{eq:score_interpolation}
    \hat{x}_0^{\mathrm{emb}} =
    \mathbb{E}[x_0^{\mathrm{emb}} \mid x_t] =
    \sum_{k=0}^{C_j-1} p_\theta(k \mid x_t) \cdot e_{j,k}
\end{align}
where $p_\theta(k \mid x_t) = \mathrm{softmax}(\mathrm{logits}_j)_k$. This estimate is then used in an Euler update analogous to that for numerical features. After sampling is complete, the final embeddings $x_0^{\mathrm{emb}}$ are converted to categorical indices via nearest-neighbor lookup.

\subsection{Classifier-Free Guidance for Conditional Generation}
\label{sec:cfg}

We incorporate \gls{cfg}~\cite{ho2022classifier} to support conditional generation, which is particularly valuable for \gls{ehr} data where important clinical events (e.g., in-hospital mortality) are rare, resulting in imbalanced label distributions. \Gls{cfg} enables generating \glspl{ehr} conditioned on a target label $y$, facilitating data augmentation for underrepresented classes.

During training, the label information $y$ is replaced with a zero vector with probability $p_{\mathrm{drop}}$, enabling the model to learn both conditional and unconditional predictions. During sampling, conditional and unconditional predictions are interpolated:
\begin{align}
    \label{eq:cfg_num}
    \tilde{x}_0^{\mathrm{num}} =
    (1 + w_{\mathrm{num}})\,\hat{x}_{0,c}^{\mathrm{num}} -
    w_{\mathrm{num}}\,\hat{x}_{0,u}^{\mathrm{num}}
\end{align}
\begin{align}
    \label{eq:cfg_cat}
    \widetilde{\mathrm{logits}} =
    (1 + w_{\mathrm{cat}})\,\mathrm{logits}_c -
    w_{\mathrm{cat}}\,\mathrm{logits}_u
\end{align}
where $\hat{x}_{0,c}^{\mathrm{num}}$ and $\mathrm{logits}_c$ are conditional predictions (with label $y$), $\hat{x}_{0,u}^{\mathrm{num}}$ and $\mathrm{logits}_u$ are unconditional predictions (with $y$ replaced by $\emptyset$), and $w_{\mathrm{num}}$, $w_{\mathrm{cat}}$ are guidance strengths for numerical and categorical features, respectively.

\section{Experiments}
\label{sec:experiments}

\subsection{Datasets}
\label{sec:datasets}

We evaluate our method on two large-scale publicly available \gls{icu} datasets. Table~\ref{tab:datasets} summarizes the dataset specifications. Both datasets are de-identified and publicly available, with access granted through the respective data use agreements.

\begin{table}[t]
    \centering
    \caption{Dataset Specifications}
    \label{tab:datasets}
    \begin{tabular}{lcc}
        \toprule
        & MIMIC-III & eICU \\
        \midrule
        Training samples & 20{,}920 & 48{,}849 \\
        Validation samples & 5{,}230 & 16{,}269 \\
        Numerical features & 7 & 4 \\
        Categorical features & 7 & 4 \\
        Sequence length & 25 & 272 \\
        Time interval & 1 hour & 5 minutes \\
        \bottomrule
    \end{tabular}
\end{table}

\textbf{MIMIC-III}~\cite{johnson_pollard_mark_2016_mimic,johnson_et_al_2016_mimic_iii,goldberger_et_al_2000_physionet}: Following the preprocessing of TimeDiff~\cite{tian2024reliable}, we extract seven physiological variables (body temperature, systolic blood pressure, diastolic blood pressure, mean arterial pressure, heart rate, respiratory rate, and oxygen saturation) at 1-hour intervals from the first 24 hours of \gls{icu} admission. Missingness masks for each variable and an in-hospital mortality flag are appended as categorical features. Missing values are imputed hierarchically using last observation carried forward, per-patient mean, and dataset-wide mean, in that order.

\textbf{eICU}~\cite{pollard2018eicu}: Similarly following TimeDiff's preprocessing, we extract four physiological variables, namely heart rate, respiratory rate, oxygen saturation, and mean arterial pressure, at 5-minute intervals from the first 24 hours. Missingness masks and a mortality flag are appended. Only patients with continuously available observations are included.

\subsection{Baselines}
\label{sec:baselines}

We primarily compare with TimeDiff~\cite{tian2024reliable}, a \gls{ddpm}-based diffusion model for mixed-type temporal \glspl{ehr}, in two variants, namely \textbf{TimeDiff-GRU} (original \gls{bigru} backbone) and \textbf{TimeDiff-Trans} (backbone replaced by a Transformer). For our method, we evaluate three generation modes, including \textbf{Ours uncond} (unconditional generation), \textbf{Ours CFG-comb} (conditional generation with \gls{cfg}, combining samples according to the original label distribution), and \textbf{Ours CFG-bal} (conditional generation with balanced sampling across labels).

\subsection{Evaluation Metrics}
\label{sec:metrics}

We evaluate synthetic data quality from three complementary perspectives:

\subsubsection{Downstream task performance}

We assess whether synthetic data can serve as a substitute for real data in training machine learning models. Binary classification models are trained on synthetic data and tested on held-out real data (\gls{tstr}). We use three classifiers, including bidirectional \gls{lstm}, Transformer, and hybrid CNN-\gls{lstm}, and report the mean \gls{auc} across classifiers and five random seeds. The real-data baseline (\gls{trtr}) serves as the upper bound.

\subsubsection{Fidelity}
We evaluate how well synthetic data preserves the statistical properties of real data. For numerical features, we compute \gls{mmd} for distributional distance, \gls{corr-mae} for inter-feature dependency preservation, \gls{acf-mse} for temporal pattern preservation, and \gls{dtw} for time-series shape similarity. For categorical features, we compute \gls{tvd} for marginal distribution fidelity, and \gls{trans-dist} for temporal transition pattern preservation.

\subsubsection{Discriminability}

The \gls{c2st} trains a binary classifier to discriminate real from synthetic data. We use three discriminators, including logistic regression, \gls{mlp}, and \gls{lstm}, and report the mean \gls{auc} over five random seeds; values closer to 0.5 indicate less distinguishable synthetic data.

\subsection{Implementation Details}
\label{sec:implementation}

Our model is trained using the Adam optimizer with a learning rate of 0.001 and a batch size of 4,096 for 8,000 epochs. Exponential moving average (EMA) with a decay rate of 0.997 is maintained for both model parameters and noise schedule parameters throughout training~\cite{karras2022elucidating}. For sampling, we use the checkpoint with the lowest EMA-based loss and generate data using the first-order Euler method with 50 steps. EDM preconditioning parameters are set to $\sigma_{\min} = 0.002$ and $\sigma_{\mathrm{data}} = 0.5$, with $\sigma_{\max}^{\mathrm{num}} = 80$ for numerical features and $\sigma_{\max}^{\mathrm{emb}} = 100$ for categorical embeddings~\cite{mueller2025continuous}. The \gls{bigru} backbone consists of 3 layers with a hidden dimension of 64. The categorical embedding dimension is $d = 16$. The loss weights are set to $\lambda_{\mathrm{num}} = \lambda_{\mathrm{emb}} = 1.0$. The global noise schedule parameter is initialized to $\rho_{\mathrm{global}} = 1.0$ for numerical features and $\rho_{\mathrm{global}} = 7.0$ for categorical embeddings, with feature-specific and time-specific components initialized to zero. For classifier-free guidance, the label drop probability is $p_{\mathrm{drop}} = 0.1$, and guidance weights are set to $w_{\mathrm{num}} = w_{\mathrm{cat}} = 2.0$.

For TimeDiff baselines, we use the original code\footnote{\url{https://github.com/MuhangTian/TimeDiff}} provided by the authors with default settings and 1,000 sampling steps~\cite{tian2024reliable}. All experiments are conducted on either a single NVIDIA TITAN RTX or a single NVIDIA RTX A6000 GPU.

\subsection{Results}
\label{sec:results}

\subsubsection{Downstream task performance}

Table~\ref{tab:tstr} presents the downstream task results using all features. Our method outperforms both TimeDiff variants on both datasets, approaching the real-data upper bound (\gls{trtr}) most closely. On MIMIC-III, Ours CFG-bal achieves the best mean \gls{auc} of 0.751, while Ours uncond achieves 0.708 on eICU. These results are achieved with only 50 sampling steps, compared to 1{,}000 steps for TimeDiff. Since both methods use a \gls{bigru} backbone with comparable per-step cost, the continuous-time formulation with unified Gaussian diffusion enables substantially faster and higher-quality generation.

On MIMIC-III, \gls{cfg} improves performance over unconditional generation, with CFG-bal outperforming CFG-comb. This is consistent with the role of balanced sampling in addressing the imbalanced label distribution (mortality rate $\approx$8\%). On eICU, however, unconditional generation performs best. We attribute this to the much longer sequence length (272 vs. 25), which dilutes the relative importance of the label that is shared across all time steps.

\begin{table}[tbp]
    \centering
    \caption{Downstream Task \gls{auc} (Mean of 3 Classifiers, $\uparrow$)}
    \label{tab:tstr}
    \begin{tabular}{lcc}
        \toprule
        Model & MIMIC-III & eICU \\
        \midrule
        \gls{trtr} & $0.770 \pm 0.005$ & $0.747 \pm 0.003$ \\
        \midrule
        TimeDiff-GRU & $0.663 \pm 0.008$ & $0.607 \pm 0.010$ \\
        TimeDiff-Trans & $0.689 \pm 0.011$ & $0.642 \pm 0.008$ \\
        \midrule
        Ours uncond & $0.729 \pm 0.006$ & $\mathbf{0.708} \pm 0.004$ \\
        Ours CFG-comb & $0.743 \pm 0.004$ & $0.688 \pm 0.005$ \\
        Ours CFG-bal & $\mathbf{0.751} \pm 0.006$ & $0.689 \pm 0.006$ \\
        \bottomrule
    \end{tabular}
\end{table}

\subsubsection{Fidelity}

Table~\ref{tab:fidelity} presents fidelity metrics for both numerical and categorical features. CFG-bal is excluded as it intentionally distorts the label distribution. Our method shows substantial improvements in most numerical metrics on both datasets. On eICU, Ours uncond achieves a 77\% reduction in \gls{mmd} (0.046 vs. 0.194--0.196), 63\% reduction in \gls{corr-mae}, and 95\% reduction in \gls{acf-mse} compared to both TimeDiff variants. On MIMIC-III, our method achieves the best \gls{corr-mae} and \gls{acf-mse}, with competitive \gls{mmd}. For \gls{dtw}, TimeDiff-GRU achieves the lowest values, as \gls{dtw} measures pairwise shape similarity and is more sensitive to sample-level variation, suggesting that our method better preserves distributional properties at the population level.

For categorical features, TimeDiff-Trans achieves the lowest \gls{tvd} on both datasets. \Gls{cfg} consistently improves both \gls{tvd} and \gls{trans-dist} for our method. The continuous embedding approach prioritizes cross-feature relationships over exact marginal reproduction, which may explain the \gls{tvd} gap while yielding substantially better downstream task performance (Table~\ref{tab:tstr}).

\begin{table*}[tbp]
    \centering
    \caption{Fidelity Metrics ($\downarrow$). Best results per dataset are \textbf{bolded}.}
    \label{tab:fidelity}
    \begin{tabular}{@{} ll cccc cc @{}}
        \toprule
        & & \multicolumn{4}{c}{Numerical} & \multicolumn{2}{c}{Categorical} \\
        \cmidrule(lr){3-6} \cmidrule(lr){7-8}
        Dataset & Method
            & \gls{mmd}
            & \gls{corr-mae}
            & \gls{acf-mse}
            & \gls{dtw}
            & \gls{tvd}
            & \gls{trans-dist} \\
        \midrule
        \multirow{4}{*}{MIMIC-III}
            & TimeDiff-GRU
                & 0.0874 & 0.0237 & 0.00523 & $\mathbf{45.9}$  & 0.036 & 0.225 \\
            & TimeDiff-Trans
                & 0.0373 & 0.0129 & 0.00477 & 48.8              & $\mathbf{0.002}$ & $\mathbf{0.008}$ \\
        \addlinespace[2pt]
            & Ours uncond
                & 0.0498 & $\mathbf{0.0100}$ & $\mathbf{0.00025}$ & 49.5 & 0.027 & 0.047 \\
            & Ours CFG-comb
                & $\mathbf{0.0356}$ & 0.0270 & 0.00040 & 50.2    & 0.021 & 0.035 \\
        \midrule
        \multirow{4}{*}{eICU}
            & TimeDiff-GRU
                & 0.1955 & 0.0365 & 0.00560 & $\mathbf{114.3}$ & 0.058 & 0.365 \\
            & TimeDiff-Trans
                & 0.1941 & 0.0461 & 0.00311 & 168.5             & $\mathbf{0.012}$ & 0.099 \\
        \addlinespace[2pt]
            & Ours uncond
                & $\mathbf{0.0456}$ & $\mathbf{0.0136}$ & $\mathbf{0.00027}$ & 135.4 & 0.040 & 0.100 \\
            & Ours CFG-comb
                & 0.0711 & 0.0564 & 0.00192 & 150.8             & 0.022 & $\mathbf{0.074}$ \\
        \bottomrule
    \end{tabular}
\end{table*}

\subsubsection{Discriminability}

Table~\ref{tab:c2st} presents \gls{c2st} results. Our method achieves \gls{auc} values substantially closer to 0.5 across all discriminators and datasets, indicating that the generated data is significantly harder to distinguish from real data.

For the simplest linear model (logistic regression), our method yields \gls{auc} close to 0.5, while TimeDiff baselines show higher \gls{auc} particularly on eICU (0.735 for GRU, 0.979 for Trans). With the \gls{mlp} discriminator, TimeDiff-GRU achieves \gls{auc} above 0.96 on both datasets, indicating near-perfect discrimination, whereas our method remains around 0.59. With the \gls{lstm} discriminator, which captures temporal patterns, our method achieves the lowest \gls{auc} values (0.516 on MIMIC-III, 0.510 on eICU), suggesting that the continuous-time diffusion model effectively reproduces temporal statistical properties.

The remaining gap with the \gls{mlp} discriminator ($\approx$0.59) suggests room for improvement in capturing complex nonlinear inter-feature dependencies.

\begin{table}[tbp]
    \centering
    \caption{Discriminability (\gls{c2st}) \gls{auc} (closer to 0.5 is better). Best results per dataset are \textbf{bolded}.}
    \label{tab:c2st}
    \resizebox{\columnwidth}{!}{%
    \begin{tabular}{@{} ll ccc @{}}
        \toprule
        Dataset & Method & Logistic & \gls{mlp} & \gls{lstm} \\
        \midrule
        \multirow{4}{*}{MIMIC-III}
            & TimeDiff-GRU
                & $.575 \pm .005$ & $.962 \pm .005$ & $.951 \pm .006$ \\
            & TimeDiff-Trans
                & $.582 \pm .005$ & $.885 \pm .020$ & $.540 \pm .006$ \\
        \addlinespace[2pt]
            & Ours uncond
                & $\mathbf{.536} \pm .004$ & $\mathbf{.593} \pm .005$ & $\mathbf{.516} \pm .004$ \\
            & Ours CFG-comb
                & $.554 \pm .002$ & $.617 \pm .005$ & $.526 \pm .005$ \\
        \midrule
        \multirow{4}{*}{eICU}
            & TimeDiff-GRU
                & $.735 \pm .003$ & $.968 \pm .002$ & $.916 \pm .011$ \\
            & TimeDiff-Trans
                & $.979 \pm .001$ & $.988 \pm .001$ & $.886 \pm .033$ \\
        \addlinespace[2pt]
            & Ours uncond
                & $\mathbf{.530} \pm .005$ & $\mathbf{.584} \pm .004$ & $\mathbf{.510} \pm .004$ \\
            & Ours CFG-comb
                & $.568 \pm .007$ & $.626 \pm .006$ & $.570 \pm .003$ \\
        \bottomrule
    \end{tabular}%
    }
\end{table}

\subsection{Ablation Studies}
\label{sec:ablation}

We conduct ablation studies on the MIMIC-III dataset to investigate the contribution of each component, using downstream task \gls{auc} as the primary metric.

\subsubsection{Categorical Variable Modeling: Mask vs. Embedding}

\begin{table}[tbp]
    \centering
    \caption{Ablation: Mask Diffusion vs. Continuous Embedding\\
    Downstream Task \gls{auc} (Mean of 3 Classifiers, $\uparrow$)}
    \label{tab:ablation_mask_emb}
    \begin{tabular}{lcc}
        \toprule
        Mode & Ours-Mask & Ours-Emb \\
        \midrule
        uncond & $0.711 \pm 0.005$ & $0.729 \pm 0.006$ \\
        CFG-comb & $0.671 \pm 0.007$ & $0.743 \pm 0.004$ \\
        CFG-bal & $0.678 \pm 0.009$ & $\mathbf{0.751} \pm 0.006$ \\
        \bottomrule
    \end{tabular}
\end{table}

Table~\ref{tab:ablation_mask_emb} compares two categorical variable modeling approaches: mask diffusion (Ours-Mask, as in the original TabDiff) and continuous embedding with Gaussian diffusion (Ours-Emb, our proposed approach). Ours-Emb outperforms Ours-Mask across all evaluation models and generation modes. The improvement is particularly pronounced with \gls{cfg}, where Ours-Emb achieves an \gls{auc} of 0.751 (CFG-bal) compared to 0.678 for Ours-Mask. This confirms the effectiveness of unified Gaussian diffusion: because all features share the same continuous diffusion space, cross-feature dependencies are maintained during conditional generation, whereas mask diffusion processes categorical features independently from numerical features.

\subsubsection{Learnable vs. Fixed Noise Schedule}

\begin{table}[tbp]
    \centering
    \caption{Ablation: Learnable vs. Fixed Noise Schedule\\
    Downstream Task \gls{auc} (Mean of 3 Classifiers, $\uparrow$)}
    \label{tab:ablation_schedule_tstr}
    \small
    \begin{tabular}{lcc}
        \toprule
        Mode & Learnable & Fixed \\
        \midrule
        uncond & $\mathbf{0.729} \pm 0.006$ & $0.707 \pm 0.009$ \\
        CFG-comb & $\mathbf{0.743} \pm 0.004$ & $0.742 \pm 0.006$ \\
        CFG-bal & $\mathbf{0.751} \pm 0.006$ & $0.745 \pm 0.006$ \\
        \bottomrule
    \end{tabular}
\end{table}

Table~\ref{tab:ablation_schedule_tstr} compares the factorized learnable noise schedule (proposed) with a fixed schedule (where the shape parameter $\rho$ is not learned and no temporal adaptation is applied). Both configurations share the same noise range ($\sigma_{\min}$, $\sigma_{\max}$) following EDM~\cite{karras2022elucidating} and \gls{cdtd}~\cite{mueller2025continuous}; only the $\rho$ parameter controlling the interpolation curve shape is learned.

As noted by Karras et al.~\cite{karras2022elucidating}, the noise range dominates generation quality in the continuous-time limit. However, with finite sampling steps (50 in our case), $\rho$ determines the step allocation: $\rho > 1$ allocates more steps to low-noise regions, and $\rho < 1$ to high-noise regions. Learning $\rho$ per feature and time step thus enables automatic adaptation of step allocation to each feature's difficulty.

The learnable schedule improves downstream task performance (Table~\ref{tab:ablation_schedule_tstr}), with the largest gain in the unconditional setting (0.729 vs. 0.707). Fidelity metrics also favor the learnable schedule in most indicators, while \gls{c2st} results show marginal trade-offs (see the supplementary material for details).

\subsubsection{Data Augmentation with Real Data}

Table~\ref{tab:augmentation} evaluates data augmentation on MIMIC-III. In the upper half, limited real data is supplemented with synthetic data to match the full dataset size (20{,}920 samples) using the T(S+R)TR framework. CFG-based augmentation with 50\% real data achieves a mean \gls{auc} of 0.763, reaching 99.1\% of the full real-data baseline (\gls{trtr}: 0.770). \Gls{cfg} outperforms unconditional generation across all settings, with the advantage being more pronounced at lower real data fractions. This is because \gls{cfg} explicitly conditions on labels, compensating for the minority class even when the unconditional model underrepresents rare events.

The lower half examines whether adding synthetic data to the full real dataset can improve performance. Unconditional augmentation at 100\% and 200\% achieves \glspl{auc} of 0.775 and 0.776, exceeding the baseline of 0.770, suggesting that diversity from synthetic data improves classifier generalization. However, performance declines with further augmentation, suggesting that excessive augmentation dilutes real data signals (see the supplementary material for detailed results across augmentation scales).

\begin{table}[tbp]
    \centering
    \caption{Data Augmentation on MIMIC-III\\
    Downstream Task \gls{auc} (Mean of 3 Classifiers, $\uparrow$)}
    \label{tab:augmentation}
    \small
    \begin{tabular}{lcc}
        \toprule
        Setting & uncond & CFG-comb \\
        \midrule
        \gls{trtr} (100\% Real) & \multicolumn{2}{c}{$0.770$} \\
        \midrule
        10\% Real + 90\% Synth & $0.693 \pm 0.005$ & $\mathbf{0.726} \pm 0.002$ \\
        25\% Real + 75\% Synth & $0.681 \pm 0.010$ & $\mathbf{0.734} \pm 0.003$ \\
        50\% Real + 50\% Synth & $0.753 \pm 0.005$ & $\mathbf{0.763} \pm 0.003$ \\
        \midrule
        100\% Real $+$ 100\% Synth & $0.775 \pm 0.005$ & $0.768 \pm 0.004$ \\
        100\% Real $+$ 200\% Synth & $\mathbf{0.776} \pm 0.004$ & $0.768 \pm 0.005$ \\
        \bottomrule
    \end{tabular}
\end{table}

\section{Conclusion}
\label{sec:conclusion}

We proposed a continuous-time diffusion framework for generating mixed-type time-series electronic health records. By extending TabDiff with a \gls{bigru} temporal backbone, unified Gaussian diffusion via continuous categorical embeddings, and factorized learnable noise schedules, our method achieves state-of-the-art performance on MIMIC-III and eICU datasets in downstream task utility, distribution fidelity, and discriminability, while requiring only 50 sampling steps compared to 1{,}000 for discrete-time baselines. The ablation study confirms that unified Gaussian diffusion is the primary contributor, and classifier-free guidance further enables effective conditional generation for data augmentation in class-imbalanced clinical scenarios.

Our method shows room for improvement in categorical marginal distribution fidelity and complex nonlinear inter-feature dependencies. Future work includes addressing these limitations, validating on additional \gls{ehr} datasets such as HiRID~\cite{hyland2020early}, and exploring applications to other temporal clinical data domains.

\section*{Acknowledgment}

The authors would like to express their sincere gratitude to Prof. Dr. Prof. h.c. Andreas Dengel for his understanding and support in conducting this research.

\section*{References}
\bibliographystyle{IEEEtran}
\bibliography{reference}

\end{document}